\documentclass[onecolumn]{ieeetran}
\usepackage{cite}
\usepackage{amsmath,amssymb,amsfonts}
\usepackage{algorithmic}
\usepackage{graphicx}
\usepackage{textcomp}
\begin{document}
%\history{Date of publication xxxx 00, 0000, date of current version xxxx 00, 0000.}
%\doi{10.1109/ACCESS.2017.DOI}

\title{Lossless Compression of Point Cloud Sequences Using Sequence Optimized CNN Models}
\author{\uppercase{Emre C. Kaya}, \uppercase{and}
\uppercase{Ioan Tabus %\IEEEmembership{Senior Member, IEEE} 
}}
%\address{Computing Sciences Unit, Tampere University, 33720 Tampere, Finland }

\markboth
{E. C. Kaya, I. Tabus: Lossless Compression of Point Cloud Sequences using Sequence Optimized CNN Models}
{E. C. Kaya, I. Tabus: Lossless Compression of Point Cloud Sequences using Sequence Optimized CNN Models}

%\corresp{Corresponding author: Emre C. Kaya (e-mail: emre.kaya@tuni.fi).}
\maketitle

\begin{abstract}
We propose a new paradigm for encoding the geometry of point cloud sequences, where the convolutional neural network (CNN) which estimates the encoding distributions is optimized on several frames of the sequence to be compressed. We adopt lightweight CNN structures, we perform training as part of the encoding process, and the CNN parameters are transmitted as part of the bitstream. The newly proposed encoding scheme operates on the octree representation for each point cloud, encoding consecutively each octree resolution layer. At every octree resolution layer, the voxel grid is traversed section-by-section (each section being perpendicular to a selected coordinate axis) and in each section the occupancies of groups of two-by-two voxels are encoded at once, in a single arithmetic coding operation. A context for the conditional encoding distribution is defined for each two-by-two group of voxels, based on the information available about the occupancy of neighbor voxels in the current and lower resolution layers of the octree. The CNN estimates the probability distributions of occupancy patterns of all voxel groups  from one section in four phases. In each new phase the contexts are updated with the occupancies encoded in the previous phase, and each phase estimates the probabilities in parallel, providing a reasonable trade-off between the parallelism of processing and the informativeness of the contexts. The CNN training time is comparable to the time spent in the remaining encoding steps, leading to competitive overall encoding times. Bitrates and encoding-decoding times compare favorably with those of recently published compression schemes.
\end{abstract}

%\begin{keywords}
%Convolutional neural networks, lossless geometry compression, octree coding, point cloud compression
%\end{keywords}

%\titlepgskip=-15pt

\section{Introduction}
\label{sec:introduction}
The compression of the voxelized point clouds became a hot research topic recently, owing to the need of developing immersive technologies, underlined e.g., in the programs launched by MPEG \cite{mpeg} and JPEG \cite{jpeg} standardization bodies, which  already resulted in two well-engineered standards, V-PCC \cite{vpcc} and G-PCC \cite{gpcc2} (having the test model TMC13 \cite{gpcc}). 
The scientific literature has  witnessed a strong interest in improving the compression performance of G-PCC, with many contributions in recent years, e.g., \cite{peixoto2020intra,s4d,wang2021multiscale,comprehensive,9154552,nguyen2021learning,ramalho2021silhouette,BVL,quach2019learning,garcia2017context,octsqueeze,nnoc,garcia2019geometry,de2018distance,que2021voxelcontext,wang2021lossy,wen2020lossy}. These methods differ in several aspects such as the representation used for the point cloud (e.g., octree \cite{garcia2017context,octsqueeze,nnoc,garcia2019geometry,de2018distance,que2021voxelcontext}, dyadic decomposition \cite{peixoto2020intra,s4d,ramalho2021silhouette} or projections onto 2D planes \cite{tzamarias2021compression,BVL}), the selection of the context used for the conditional probability model for arithmetic coding and also the way symbols to be encoded are defined.  

The probability model can be based on adaptively maintained counts for various contexts \cite{gpcc2,s4d}, or on a NN model \cite{nguyen2021learning,nnoc}, having a binary context at the input and the probability mass function of the symbol at the output.

\begin{figure*}

\centerline{\includegraphics[width=\columnwidth]{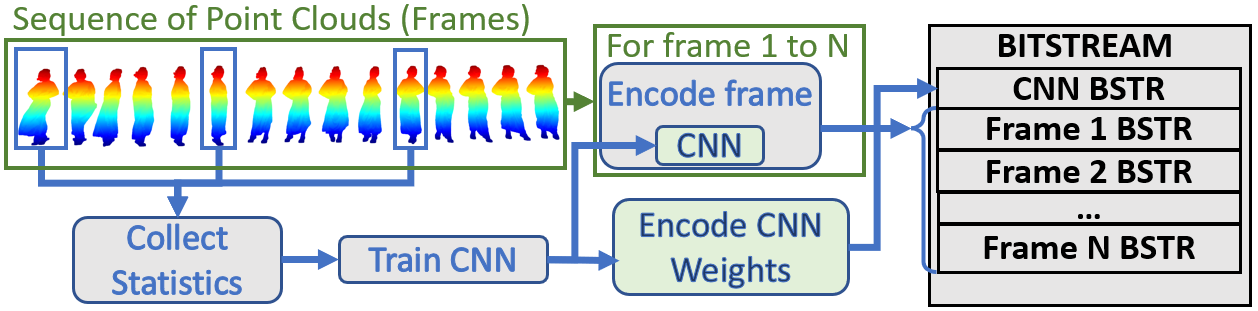}}
\caption{Overview of the proposed scheme for encoding point cloud sequences. }
\label{fig:overview}
\end{figure*} 

In the last few years, machine learning approaches using neural networks were proved to be successful for both lossy \cite{wang2021lossy,quach2019learning,wen2020lossy,que2021voxelcontext,octsqueeze} and lossless \cite{nnoc,nguyen2021learning,que2021voxelcontext} point cloud geometry compression. In \cite{quach2019learning}, an auto-encoder architecture involving 3D convolutional layers is employed to generate a latent representation of the point cloud which is further compressed using range coding. In \cite{wang2021lossy}, a variational auto-encoder model is employed in an end-to-end learning scheme. Wen et al. \cite{wen2020lossy} perform adaptive octree-based decomposition of the point cloud prior to encoding with a multilayer perceptron based end-to-end learned analysis-synthesis architecture.

VoxelDNN \cite{nguyen2021learning} uses 3D masked convolutional filters to enforce the causality of the 3D context from which the occupancy of  $64\times64\times64$ blocks of voxels are estimated. VoxelContext-Net \cite{que2021voxelcontext} employs an octree-based deep entropy model for both dynamic and static LIDAR point clouds. NNOC \cite{nnoc} operates on the octree representation where hybrid contexts are formed by combining the information from two consecutive resolution levels.

As a main difference from the previous approaches that utilize neural networks, we optimize a specific neural network model for the sequence to be encoded. In the proposed scheme, the optimization (training) of the network is a part of the encoding stage and the optimized CNN model parameters are transmitted as a header to the decoder, being then used for decoding any point cloud from the sequence.

We consider here two distinct paradigms for using a neural network as coding probability model: in the first, which we dub {\em Generically Trained Model} (GTM), some generic training set is selected and is used for optimizing a NN model to be used by both encoder and decoder, for all the compression tasks that will be required in the future \cite{nguyen2021learning,nnoc,wang2021multiscale,quach2019learning,octsqueeze,que2021voxelcontext,wang2021lossy,wen2020lossy,msvoxeldnn,AdaptiveDeep}. The compression performance of the methods involving this approach depends on the training set. The model is an integral part of the encoding program, and an identical copy is assumed to exist in the decoding program, so the NN model is not considered as a part of the encoded data. Hence, the model can be made as complex as needed, since its size does not contribute to the size of the bitstream. Also, the training time of the model is not accounted for in the encoding time, although it can be rather important, of the order of hours. 
  
We propose here a second paradigm, {\em Specifically Trained Model} (STM), where the NN model to be used for the compression of a sequence of point clouds (PC) is optimized for the sequence and it is transmitted as a part of the encoded stream. One cannot simply adhere to this paradigm by selecting the same model structure as in GTM case and train it on the sequence, because the complexity-compression ratio trade-off needs to be different: the model needs to be trained fast and the cost of transmitting the model parameters needs to be sufficiently low. Using the specifically trained models paradigm as a basis for the compression of a point cloud sequence has the advantage that the NN model that generates the coding distribution at each context can be trained to match very closely the real distributions found in the point clouds that form the sequence, even with a less complex NN structure than in the general trained model case. Another advantage is that the point clouds in a sequence are similar one to another from the point of view of their real probability distributions at the contexts, whereas in the case of a general trained model the distributions generated by the model at each context can differ quite much from the distributions learned from the generic training data.

In this paper we propose a solution which belongs to STM paradigm and has the following main features: it uses the octree representation \cite{meagher1982geometric}; encodes the occupancy of groups of $2 \times 2$ voxels at each arithmetic encoding operation; uses a context based on occupancies of lower resolution voxels (translated as candidate voxels at the current resolution) and on sets of voxels already encoded at the current resolution; computes the probability using a CNN model having at the input a multivalued context (which becomes gradually more relevant, along four phases). 

The structure of the paper is as follows. We introduce our method in Section II, we present experimental results in Section III and we present conclusions in Section IV.

\begin{figure*}

\centerline{\includegraphics[width=\linewidth]{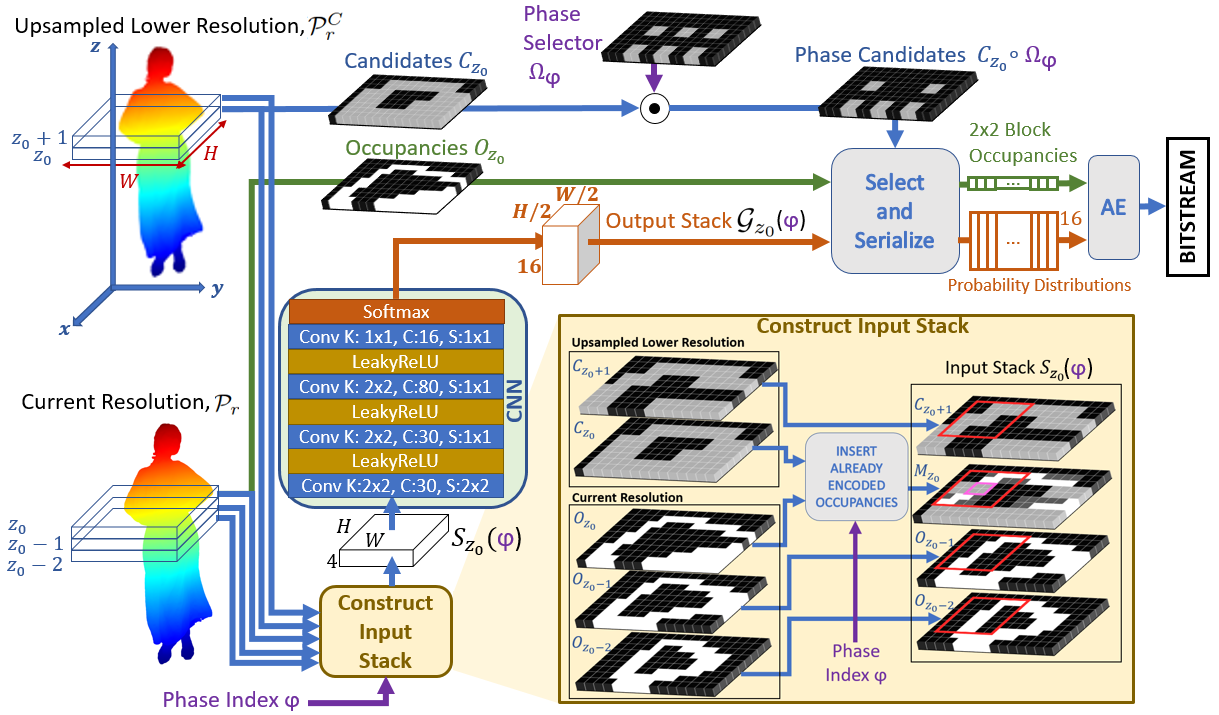}}
\caption{Encoding of the occupancy image $O_{z_0}$ (the section $z=z_0$ through the point cloud ${\cal P}_r$ at resolution $r$) at phase $\varphi$. In the CNN block, K denotes the 2D kernel size, C denotes the number of output channels and S is the 2D stride. }
\label{fig:encframe}
\end{figure*} 

\section{Proposed Method}

\label{sec:description}

We consider encoding the geometry of the point clouds forming a sequence, each point cloud being called a frame. An overview of the proposed sequence encoding scheme is provided in Fig. \ref{fig:overview}. The encoding scheme consists of three stages. In the first stage, we collect contexts and corresponding histograms as defined in Section II-A, from a small number of equidistant frames from the sequence. In the second stage, we train a fixed structure CNN model using the contexts and corresponding statistics which were collected in the first stage. The resulting optimal weights of the CNN are losslessly transmitted consuming 32 bits for each weight. In the final stage, we encode the entire sequence using the same CNN for each frame. The frames can be encoded independently one from another, hence they can be decoded in any order, independently one from another, hence resulting in a random access at the point clouds of the sequence. We note that the methods using interframe coding do not  posses this random access property.

\subsection{Encoding of frames}
\label{sec:encframes}

At each frame, the octree representation (in breadth-first order) of the point cloud is processed iteratively on the resolution $r$,  starting from the  resolution $r=2$ (octree depth level) and eventually reaching the original resolution. First, the PC at $r=2$ is written to the bitstream in 64 bits where each bit represents the occupancy of a voxel. 

The encoding steps at resolutions higher than 2 are illustrated in Fig. \ref{fig:encframe}. At each resolution level $r$, we have available at both encoder and decoder the point cloud ${\cal P}_{r-1}$, and we create from it an upsampled version  ${\cal P}^C_{r}$, by splitting each occupied voxel of ${\cal P}_{r-1}$ into eight candidate voxels in ${\cal P}^C_{r}$, having now resolution of $r$ bits per dimension. 
% the reconstruction of the one step lower resolution ${\cal P}_{r-1}$ by 2 (along all 3 dimensions), obtaining a point cloud of candidate locations ${\cal P}^C_{r}$. 
 For encoding the  current resolution ${\cal P}_{r}$ (the true occupancies), we need to encode (transmit) the occupancy status of each candidate voxel in ${\cal P}^C_{r}$. This is performed sequentially by sweeping $z_0$  along the sweeping dimension $z$, and at each $z_0$ considering the sectioning by the plane $z=z_0$ of the point cloud ${\cal P}^C_{r}$, and for each candidate voxel transmitting the occupancy status.  

To use suggestive geometric interpretation, when encoding the voxels at the section $z=z_0$ in the current resolution $r$ of the PC ${\cal P}_{r}$, we refer to the plane in $x$ and $y$ coordinates, as a binary $(W\times H)$-occupancy image $O_{z_0}$. The section $z=z_0$ through the upsampled candidate point cloud ${\cal P}^C_r$, is called the $(W\times H)$-candidate image, $C_{z_0}$. During the encoding (and the decoding) process, the reconstructed occupancy image $R_{z_0}(\varphi)$ (a binary $(W\times H)$-image)  is maintained for each of the four phases $\varphi = 1,2,3,$  and $4$, as described in Subsection \ref{sec:fourphases}. When encoding the pixels of $O_{z_0}$, the planes $O_{z_0-1}$ and $O_{z_0-2}$ were already encoded, and we rely, when building conditioning distributions by the CNN, on the most relevant available information, namely the images $C_{z_0+1},C_{z_0},O_{z_0-1},O_{z_0-2}$, and also for the information in the current plane $z=z_0$ reconstructed as $R_{z_0}(\varphi)$. The information from the candidate image $C_{z_0}$ and reconstructed image $R_{z_0}$ is combined into a four-level mixing image, $M_{z_0}(\varphi)$  as described in Subsection \ref{sec:fourphases}. 
%EMRE: should we tell Rz0 is reconstruction of 0z0 somewhere?
The relevant information for encoding $O_{z_0}$ is finally gathered in the $4\times W\times H$ array called Input Stack $S_{z_0}(\varphi)$, (constructed by the block “Construct Input Stack” in Fig. \ref{fig:encframe}), 
which is input to a CNN for estimating probabilities for occupancy of the voxels at $z=z_0$. 

We encode the pixels of $O_{z_0}$ in groups of $2\times2$ pixels and for each $(2\times2)-$block $$B_{m,n} =  \left[\begin{array}{cc}O_{z_0}(2m,2n)&O_{z_0}(2m,2n+1)\\ O_{z_0}(2m+1,2n)& O_{z_0}(2m+1,2n+1)\end{array}\right],$$ we define the occupancy pattern of the block as $Q_{m,n} = O_{z_0}(2m,2n)+2O_{z_0}(2m,2n+1)+2^2O_{z_0}(2m+1,2n)+ 2^3O_{z_0}(2m+1,2n+1)$. We cover the entire $W\times H$ image by nonoverlapping blocks by setting $m=0,1,\ldots, W/2-1 $ and $n=0,1,\ldots, H/2-1 $ ($W$ and $H$ are enforced to be even).

\begin{figure}

\centerline{\includegraphics[width=0.7\linewidth]{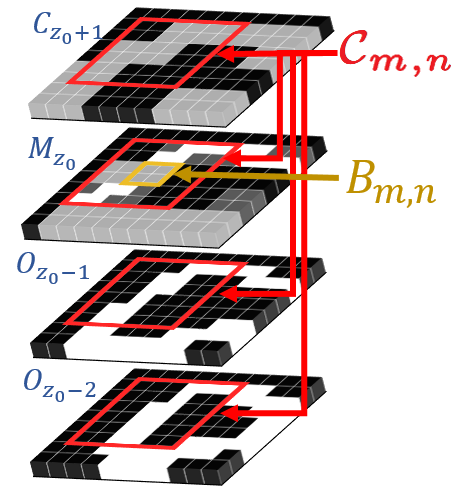}}
\caption{The context ${\cal C}_{m,n}$ (the 4x6x6 block of voxels marked in red) for computing the 16-element probability mass function $Prob(Q_{m,n}=q|{\cal C}_{m,n})$ for $q = 0,\ldots,15$, needed for encoding and decoding the occupancies of the $2\times2$ block $B_{m,n}$ marked by the yellow square. The CNN computes in parallel all these probabilities at all possible contexts. At the training stage, for each context observed in the training set, a 16-element histogram $h(q|C_{m,n})$, $q = 0,\ldots,15$, is collected and used in the Loss function from (\ref{eq:loss}).}
\label{fig:newfig}
\end{figure}

\begin{figure*}

\centerline{\includegraphics[width=\linewidth]{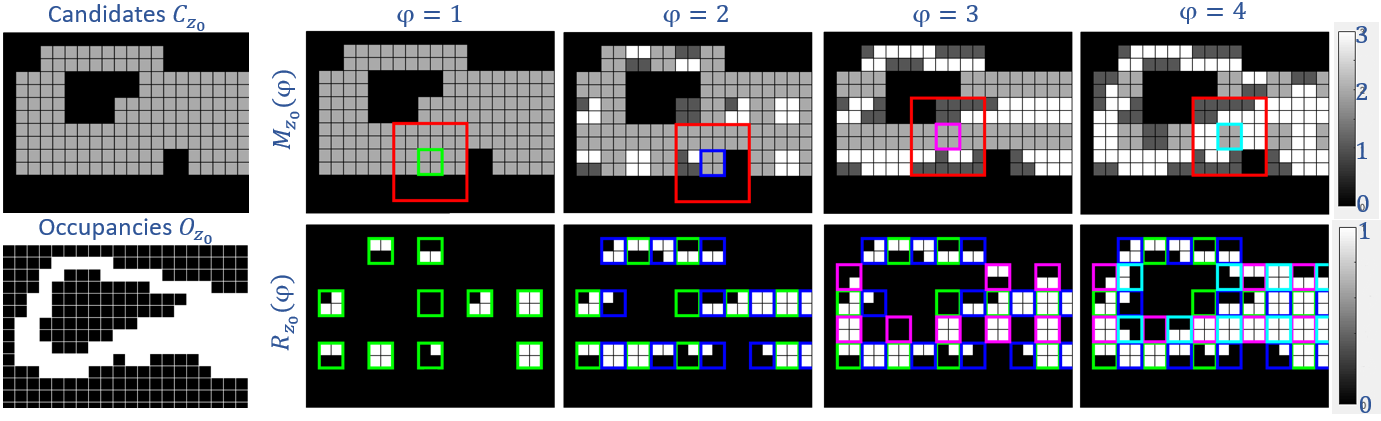}}
\caption{Inserting already encoded occupancies to construct $M_{z_0}(\varphi)$, when the candidates and occupancies images are as shown in column 1. Columns 2-5 correspond to phases 1-4. Upper row: Mixing image, $M_{z_0}(\varphi)$, to be used in input stack at phase $\varphi$. Lower row: Reconstructed image, $R_{z_0}(\varphi)$, after each phase. 
$M_{z_0}(\varphi)$ equals $2C_{z_0}$ at $\varphi=1$. The locations encoded at phase $\varphi=1$ are shown in green squares in the reconstructed $R_{z_0}(1)$. At those locations $(i,j)$ we are setting in  $M_{z_0}(2)$   the   true occupancy values, represented as $2O_{z_0}(i,j)+1$. A pixel in $M_{z_0}(\varphi)$, can have one of the four values 0,1,2,3. If it was not yet encoded, it can be 0 (if it's not a candidate, $C_{z_0}(i,j)=0$) or 2 (if it's a candidate, $C_{z_0}(i,j)=1$). If it is already encoded it can be 1 (if it's not occupied, $O_{z_0}(i,j)=0$) or 3 (if it's occupied, $O_{z_0}(i,j)=1$). At each phase, we show in $M_{z_0}(\varphi)$ one example context window as a red bounding box around its corresponding $2\times2$-block to be encoded.
 }
\label{fig:phases}
\end{figure*}

\begin{figure}

\centerline{\includegraphics[width=\columnwidth]{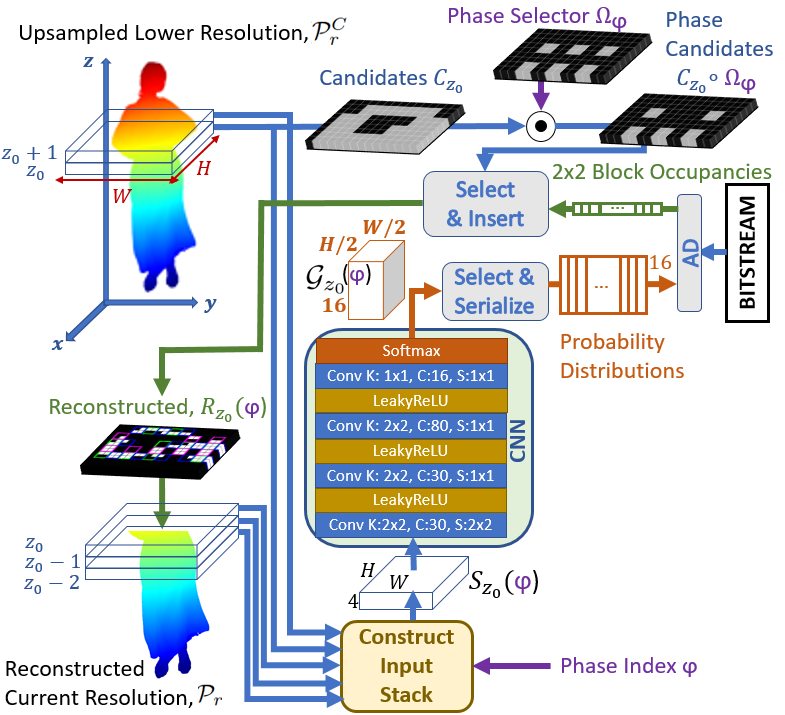}}
\caption{Decoding of the occupancy image $O_{z_0}$ (the section $z=z_0$ through the point cloud ${\cal P}_r$ at resolution $r$) at a phase $\varphi$.
The  block ``Select \& Insert'' is selecting from the image $C_{z_0}\circ \Omega_\varphi$ the locations where to save the occupancies for each decoded $2\times2-$ block. The Arithmetic Decoder (AD) decodes the serialized block occupancies for phase candidates and they are inserted into correct locations in $R_{z_0}(\varphi)$. In the next phase, $R_{z_0}(\varphi)$ is used to form the $M_{z_0}(\varphi+1)$ in the input stack $S_{z_0}(\varphi+1)$. }
\label{fig:decode}
\end{figure} 

For encoding with arithmetic coding the occupancy pattern $Q_{m,n}$ of a generic  $(2\times2)-$ block of pixels $B_{m,n}$, we utilize a probability mass function (pmf), denoted as a 16-length vector $\mathbf{G}$, specifying  at element $q$ the probability $G_q(m,n) = \text{Prob}(Q_{m,n}=q| {\cal C}_{m,n}) $, for each $q \in \{0,\ldots,15\}$, conditioned on a context ${\cal C}_{m,n}$ (see Fig. \ref{fig:newfig}). We implement the probability model by a CNN with the set of parameters $\cal W$ and having 16 output channels, which will output at every location $(m,n)$ the pmf vector denoted
\begin{equation}\mathbf{G}(m,n)= CNN_{\cal W}({\cal C}_{m,n}) = \mathbf{g}({\cal W},{\cal C}_{m,n})\label{CNNoutput}.
\end{equation}
Hence, the output of the CNN is an array of size $16 \times W/2 \times H/2$, called output stack, denoted ${\cal G}_{z_0}$. At location $(m,n)$ and channel $q$, ${\cal G}_{z_0}(q,m,n) =G_q(m,n) $. 

The conditioning is done on a context ${\cal C}_{m,n}$ which is defined by the receptive field of the CNN, i.e., by the set of pixels from the input stack $S_{z_0}$ which affect the computation of $\mathbf{G}(m,n)$. For the selected structure of the CNN shown in Fig. \ref{fig:encframe}, the receptive field ${\cal C}_{m,n}$ can be seen to be the $4 \times 6 \times 6$ subblock from the input stack $S_{z_0}$, ranging on $x$ coordinate from $2m-2$ to $2m+3$ and on $y$ coordinate from $2n-2$, to $2n+3$. In Fig. \ref{fig:encframe} and Fig. \ref{fig:newfig}, we show by red contours, the  context of $4\times 6 \times 6$ pixels  from the images $O_{z0-2}, ..., C_{z_0+1}$, corresponding to the $2\times2-$ candidate block marked in pink on the combined image $M_{z_0}(\varphi)$. 

The CNN consists of four 2D convolutional stages where convolutions operate along $W$ and $H$ dimensions and the number of input channels is four (the number of 2D images in $S_{z_0}(\varphi)$). As a nonlinear activation function at the hidden stages, we employ LeakyReLU with a constant slope of 0.01 for the negative inputs so that
\begin{equation}
\text{LeakyReLU}(x) = \max(0, x) + 0.01 * \min(0, x). 
\end{equation}

The output layer activation is chosen as softmax to ensure the output $\mathbf{G}(m,n)$ to be valid probability distributions. 

\subsection{Encoding the occupancy image $O_{z_0}$ in  four phases}

\label{sec:fourphases}

In order to improve the informativeness of the contexts, we split the encoding process in four phases such that in each phase the CNN is called once, having at the input a different input stack, $S_{z_0}(\varphi)$, which becomes more informative as we progress to the next phase. Associated to each phase $\varphi$ is a different phase selector image $\Omega_\varphi$ that selects candidates blocks to be encoded in the current phase (these are called phase candidates blocks).

Each phase selector image  $\Omega_\varphi$ is setting to 1 the pixels in a quarter of the groups of  $2\times2-$ pixels as follows.

The  elements of $\Omega_{1}$ are 1  for all  pairs  $(i,j)$  with $i= 4k,i=4k+1,j=4l,j=4l+1$ where $k,l$ are integers,  so that the element-wise product $O_{z_0}\circ \Omega_1$ ($\circ$ denotes element-
wise multiplication) forces to zero all pixels that do not belong to blocks $B_{2k,2l}$ .
% This is obtained by setting $\Omega_{1}(i,j)= 1$ for all  $i$ and $j$ of the form $i= 4k,i=4k+1,j=4\ell,j=4\ell+1$ where $k,l$ are integers. 
Similarly the  selector image $\Omega_{2}$  selects in $O_{z_0}\circ \Omega_2$ all the blocks $B_{2k,2l+1}$,  $\Omega_{3}$  selects all the blocks $B_{2k+1,2l}$, and $\Omega_{4}$  selects all the blocks $B_{2k+1,2l+1}$. 

% Subsequently, $\Omega_{2}$ shifts all pixels of $\Omega_{1}$ two positions to the left, $\Omega_{3}$ shifts all pixels of $\Omega_{1}$ two positions down, and $\Omega_{4}$ shifts all pixels of $\Omega_{3}$ two positions to the left, so that the OR function of the four $W\times H$ images $\Omega_{1},\ldots \Omega_{4}$ results in an all-ones image.

In each phase, the CNN uses the input stack, $S_{z_0}(\varphi)$ and generates pmf vectors $\mathbf{G}(m,n)= CNN_{\cal W}({\cal C}_{m,n})$ for all blocks $B_{m,n}$, with $m=0,1,\ldots W/2-1 $ and $n=0,1,\ldots H/2-1 $, hence covering all blocks of the $(W\times H)$ images.  The candidate image $C_{z_0}$ specifies which of the blocks are already know to be zero, and hence do not need to be encoded.

In the first phase, out of all $2\times2-$ blocks of $C_{z_0}$ and $O_{z_0}$, only a quarter of blocks are selected, by using $\Omega_1$, namely   from $O_{z_0}$ are selected only the blocks 
$B_{2k,2l}$, with $k=0,\ldots,W/4-1$ and $l=0,\ldots,H/4-1$. The ``Select and serialize'' block in Fig. \ref{fig:encframe} takes at the input the binary image $C_{z_0}\circ \Omega_1$, and downsamples it by 2 in both dimensions, to pick the blocks $B_{2k,2l}$ that have to be encoded. The pmf corresponding to each block is read from the pmf vector $\mathbf{G}(2k,2l)$ and is used by the arithmetic encoder.
The encoder and the decoder are now accounting that one quarter of the image $O_{z_0}$  was reconstructed, and those values can be inserted now in the input stack (this is the process marked in Fig. \ref{fig:encframe} as ``Insert already encoded occupancies'' and illustrated in detail in Fig. \ref{fig:phases}), in order to make the contexts more informative.

The phases 2,3 and 4 proceed in a similar way, after which the entire $O_{z_0}$ pixels can be reconstructed, so that $R_{z_0}=O_{z_0}$, and the algorithm moves to the processing of the next section, $z=z_0+1$.
% the CNN is computing again probabilities for all blocks, but using now the more informative contexts from the updated Input stack, $S_{z_0}(2)$. After that, again only a quarter of the blocks (and of the pixels as well), corresponding to locations where $\Omega_2(i,j) = 1$ are checked if they are also candidates, and hence the pixels of $O_{z_0}$ (in all blocks of the form $B_{2k,2l+1}$) signaled by units in $C_{z_0} \circ \Omega_2$   are  encoded and transmitted as a partial bitstream, after which both encoder and decoder prepare the Input stack for phase 3. Phases 3 and 4 proceed in the same way, after which the entire $O_{z_0}$ pixels can be reconstructed, and the algorithm moves to the processing of the next section, the image $O_{z_0+1}$.

In Fig. \ref{fig:phases}, it is shown how the candidate image $C_{z_0}$ and the occupancy image $O_{z_0}$ (or equivalently the reconstructed image $R_{z_0}$) are combined differently at each phase, resulting in the mixed $W\times H$-image $M_{z_0}(\varphi)$.
%which reflects the available information at phase $\varphi$ at section $z_0$:The candidate values for all pixels of the section, and the already reconstructed values during the previous phases,  $1,\ldots,(\varphi-1)$. 
Initially the reconstructed image $R_{z_0}(1)$ is all-zeros, then, after each phase, the occupancy at the locations where $\Omega_\varphi(i,j)=1$ is recovered: $R_{z_0}(\varphi)= R_{z_0}(\varphi-1) + O_{z_0}\circ \Omega_\varphi $.
The already processed part of the image after phase $\varphi$ is  $\Omega_\varphi^s = \Omega_1 \vee \ldots \vee \Omega_\varphi $ (element-wise OR).
The mixed image $M_{z_0}(\varphi)$ at the beginning of phase $\varphi$ is obtained as 
\begin{equation}
M_{z_0}(\varphi)= 2C_{z_0}\circ (1-\Omega_{\varphi-1}^s) +  (2R_{z_0}(\varphi)+1)\circ \Omega_{\varphi-1}^s .
\label{eq:mz0}
\end{equation}
In (\ref{eq:mz0}), the encoder can use $O_{z_0}$ instead of $R_{z_0}$.

Fig. \ref{fig:decode} shows the decoder's flow diagram, using the same probability modeling and the same CNN as the encoder. The decoder counterpart of the "Select and Serialize" block in the encoder is called the "Select and Insert" block and it inserts the decoded $2\times2$ block occupancies of phase candidates into correct locations in $R_{z_0}(\varphi)$.

\begin{table*}
\caption{Comparing Average Bitrates of SeqNOC (proposed) with Other Recent Methods}
\label{table:compares4d}
\small
\center
\begin{tabular}{|c||c|c|c|c|c||c|c|c|c|}

\hline

& \multicolumn{5}{|c||}{\textbf{Methods using models adapting to the PC or}} & \multicolumn{4}{|c|}{\textbf{Methods using models trained on}} \\

& \multicolumn{5}{|c||}{\textbf{to the sequence to be compressed}} & \multicolumn{4}{|c|}{\textbf{a generic dataset}} \\

 \cline{2-10} 

& \textbf{TMC13} &  \textbf{S3D }& \textbf{P(Full) }&  \textbf{S4DCS } & \textbf{SeqNOC} & \textbf{fNNOC} & \textbf{NNOC} & \textbf{VDNN}  & \textbf{MSVDNN}  \\

& \cite{gpcc} &  \cite{peixoto2020intra} & \cite{garcia2019geometry} &   \cite{ramalho2021silhouette} &  & \cite{nnoc} & \cite{nnoc} & \cite{nguyen2021learning} & \cite{msvoxeldnn} \\
\hline
Andrew  & 1.14 & 1.12 & 1.37 & 0.95 & \textbf{0.85}  & -  & - & - & -\\
\hline

David & 1.07 & 1.06 & 1.31&  0.94 & \textbf{0.78}  & -  & - & - & - \\
\hline

Phil & 1.17 & 1.14 & 1.42 &  1.02 & \textbf{0.86}  & 1.00 & \textit{\textbf{0.81}} & 0.92 & -\\
\hline

Ricardo & 1.09 & 1.04 & 1.34 &  0.90 & \textbf{ 0.79 }  & 0.86 & \textit{\textbf{0.68}} & 0.72  & -\\
\hline

Sarah & 1.07 & 1.07 &  1.37  &  0.92 & \textbf{0.80}  & -  &  - & - & -\\
\hline
\hline
\textbf{Average$_{\text{MVUB}}$} & 1.11 & 1.09 & 1.36  & 0.95 & \textbf{0.82}  & - & - & - & -\\
\hline
\hline
LongDress  & 1.02 & 0.95 & 1.13  & 0.88 & \textbf{0.70}  & - & - & - & -\\
\hline

 Loot & 0.96 & 0.92 & 1.02  & 0.84 & \textbf{0.66}  & 0.74 & \textit{\textbf{0.59}} & 0.64  & 0.73\\
\hline

RedAndBlack & 1.08 & 1.02 & 1.23 & 0.94 & \textbf{0.77}  & 0.87 & \textit{\textbf{0.72}} &  0.73 & 0.87 \\
\hline

Soldier & 1.03 & 0.96 & 0.85  &\textbf{ 0.65} & 0.70  & - & - & - & -\\
\hline
\hline
\textbf{Average$_{\text{8i}}$} & 1.02 & 0.96 & 1.06  &  0.83 & \textbf{0.71}  & - & - & - & -\\

\hline

\end{tabular}

\vspace{1cm}

\caption{Bitrates [bpp], Encoding Times ($t_e[s]$) and Decoding Times ($t_d[s]$) for four different configurations of the algorithm and for the method fNNOC \cite{nnoc}}
\center
% Please add the following required packages to your document preamble:
% \usepackage{multirow}
\begin{tabular}{|l|r|r|r||r|r|r||r|r|r||r|r|r||r|r|r|}
\hline
 & \multicolumn{3}{c||}{\textbf{SeqNOC (baseline)}}    & \multicolumn{3}{c||}{\textbf{SeqNOC-LM}} & \multicolumn{3}{c||}{\textbf{SeqNOC-10}} & \multicolumn{3}{c||}{\textbf{SeqNOC-SP}}  & \multicolumn{3}{c|}{\textbf{fNNOC}}   \\ \cline{2-16} 

& bpp & $t_e[s]$ & $t_d[s]$  & bpp &  $t_e[s]$ & $t_d[s]$ & bpp &  $t_e[s]$ & $t_d[s]$ & bpp &  $t_e[s]$ & $t_d[s]$ & bpp &  $t_e[s]$ & $t_d[s]$ \\ \hline
Andrew   & 0.85  &  8.4     & 7.0    &  0.87    &  7.3    & 6.8    &  \textbf{0.83} &  9.5    & 7.1    & 0.90     & \textbf{4.9} & \textbf{2.4}    & - & -  & -  \\ \hline
David    & 0.78     &  10.0  & 9.0    &  0.79     &  10.2  & 8.5    &  \textbf{0.77} &  11.9   & 9.0    & 0.78     & \textbf{6.4} & \textbf{2.6}   & - & -  & -   \\ \hline
Phil & 0.86  &  10.0 & 8.0    &  0.88    &  9.1  & 7.9    &  \textbf{0.84} &  12.5   & 8.1    & 0.88   & \textbf{6.6}    & \textbf{2.2}   &  1.00 &  42  &  46   \\ \hline
Ricardo  & 0.79  &  9.2   & 6.5    &  0.79  &  8.1   & 6.5    &  \textbf{0.77} &  11.8   & 6.6    & 0.80     & \textbf{4.6} & \textbf{2.0} &  0.86 &  40  &  43    \\ \hline
Sarah    & 0.80  &  9.2  & 8.5    &  0.82     &  8.2   & 7.9    &  \textbf{0.78} &  10.9   & 7.7    & 0.79    & \textbf{5.9}    & \textbf{2.2}    & - & -  & -   \\ \hline
\hline
\textbf{Average$_{\text{MVUB}}$}  &  0.82     &  9.4    & 7.8    &  0.83  &  8.6    & 7.5    &  \textbf{0.80} &  11.4   & 7.7    &  0.83     &  \textbf{5.7} & \textbf{2.3}  & - & -  & - \\ \hline
\hline
Longdress &  0.70     &  10.9 & 9.0    &  0.73   &  8.7 & 8.3    &  \textbf{0.69} &  14.2   & 9.1    &  0.75     &  \textbf{6.5} & \textbf{2.7}  & - & -  & - \\ \hline
Loot &  0.66  &  10.0   & 9.0    &  0.68     &  7.9   & 8.1    &  \textbf{0.64} &  16.4   & 9.0    &  0.70     &  \textbf{6.9} & \textbf{2.6} &  0.74 &  105  &  115 \\ \hline
Redandblack    &  0.77 &  9.8   & 8.5    &  0.79     &  10.2  & 8.0    &  \textbf{0.75} &  16.0 & 8.5    &  0.81     &  \textbf{7.6} & \textbf{2.5} &  0.87 &  94  &  104  \\ \hline
Soldier  &  \textbf{0.70} &  12.7   & 10.0   &  0.74    &  10.2   & 9.3    &  \textbf{0.70} &  13.4   & 10.2   &  0.77     &  \textbf{6.4} & \textbf{3.0}   & - & -  & -    \\ \hline
\hline
\textbf{Average$_{\text{8i}}$}   & 0.71 & 10.8 & 9.1    & 0.73 &  9.3    & 8.4    & \textbf{0.70} & 15.0  & 9.2  & 0.76 & \textbf{6.9} & \textbf{2.7}  & - & -  & - \\ \hline
\end{tabular}

\label{table:abl}
\end{table*}

%For each phase, the $z=z_0$ channel of the input stack $C_{z_0}$ is defined in a slightly different way since the available information in $z=z_0$ changes after each phase. Construction of $z=z_0$ channel of the input stack in four phases is depicted in Fig. \ref{fig:phases}. At each phase we encode only the candidates that are captured in a predefined grid called phase selector associated with that phase such that we encode candidates $(x,y,z_0)$ for which $x=4k_x+l_x$ and $y=4k_y+l_y$ where $k_x,k_y$ are integers and $l_x,l_y$ differ for each phase. For the $1$st phase, $l_x=0,1$ and $l_y=0,1$. For the $2$nd phase, $l_x=0,1$ and $l_y=2,3$. Similarly, $3$rd and $4$th phases have $l_x=2,3$, $l_y=0,1$ and $l_x=2,3$, $l_y=2,3$ respectively. The occupancies encoded at one phase can be used as context information in the latter phases. For that purpose, the associated channel of the input stack is updated after each phase by inserting the encoded occupancies. 

%Since the encoding of candidates at section $z=z_0$ is performed in 4 phases, after each phase the available context information at $z=z_0$ changes. Accordingly, $C_{z_0}(x,y,2)$ is updated after each phase. 

%Before the first phase, $C_{z_0}(x,y,2) = 2$ for the candidate voxels and $C_{z_0}(x,y,2) = 0$ elsewhere. After each phase, $C_{z_0}(x,y,2)$ is updated to contain either a 3 or 1 instead of a 2 for the candidates which are processed at the most recent phase. After the fourth phase, $C_{z_0}(x,y,2)$ does not contain any 2. 

\subsection{Optimization of the CNN}

The CNN is trained using data collected from a number of frames which we call the training frames. The training set consists of $(4 \times 6 \times 6)$-shaped contexts that have occured in the training frames at least once. Each context is associated with a 16-element histogram $h$ containing the number of occurences of 16 possible occupancy patterns of $2 \times 2$ candidate blocks. Training data are collected from the final resolution only.

During training, the contexts from the collected set  are fed in batches of size $N_b$ and the loss is expressed as 
\begin{equation}
Loss=  -\frac{1}{N_b}\sum_{i=1}^{N_b} \sum_{q=0}^{15} h(q|{\cal C}_i)\log_2{ g_q({\cal W},{\cal C}_{i})}, \label{eq:loss}
\end{equation}
where $h(q|{\cal C}_i)$ is the number of occurences of the $q$'th occupancy pattern in the training set for the $i$'th context, ${\cal C}_i$, of the batch and $g_q({\cal W},{\cal C}_{i})$ is the corresponding output  of the CNN, see  (\ref{CNNoutput}).

\section{Experimental Work}

We have performed experiments with sequences from Microsoft Voxelized Upper Bodies (MVUB) \cite{upperbodies} and 8i Voxelized Full Bodies \cite{8ivoxelized} datasets which are dense point cloud datasets.
MVUB sequences have 9 bits resolutions whereas 8i sequences have 10 bits resolution per dimension.

In the CNN architecture (given in Fig. \ref{fig:encframe} and \ref{fig:decode}), the total number of parameters is 15116. Each parameter is transmitted in 4 bytes leading to a model codelength of $\text{CL}_m\Tilde{=}60.5$ kB. Thus, for a frame with 500k points in a 100-frames sequence, model's contribution to bitrate is less than 1\%. Due to this small size, we did not consider in this paper the problem of entropy coding the CNN parameters, which will not affect significantly the reported achieved bitrates.

Training is performed using ADAM optimizer with batch size $N_b = 10^4$ and an initial learning rate of 0.001. Since the training time is counted as part of the encoding time, the training phase is aimed to be kept short. When no improvement in loss is observed for 20 epochs for the first time, learning rate is halved. When the same thing is observed once again, the training ends. The number of training frames is by default set to 5. In order to maximize parallelism, the sweeping axis $Oz$ is selected as the shortest dimension measured from the first frame. The algorithm is implemented using PyTorch and TorchAC \cite{mentzer2019practical} on an NVIDIA RTX 2080.

The bitrate $r$ for the sequence is measured as the average of the bitrates $r_f$ over $F=100$ consecutive frames, where the bitrate $r_f$ for a frame $f$ is measured as bits-per-point (bpp)

\begin{equation}
r_f = (\text{CL}_f + \text{CL}_{m} / F ) / n_{p,f},
\end{equation}
and $\text{CL}_f$ is the codelength for encoding the frame $f$, $\text{CL}_m$ is the codelength for encoding the CNN model and $n_{p,f}$ is the number of points in frame $f$. In addition to bitrates, we also report average encoding and decoding times per frame where encoding time per frame $t_e$ includes the time spent for collecting training data and training the CNN divided by number of frames. For the baseline model, average training time over all the 9 sequences used in our experiments (including the time spent for training data collection) is 348 seconds.

In Table \ref{table:compares4d}, we compare the bitrates $r$ obtained with the proposed SeqNOC and with some recently published algorithms. The methods on the left side of the table have to resort to optimization (our method) or to tracking adaptively the count numbers at each context, for building efficient coding probability distributions. The right part of Table \ref{table:compares4d}, under the header  ``Methods using models trained on a generic dataset'' contains methods following the GTM paradigm \cite{nguyen2021learning,nnoc,msvoxeldnn}. We note that for the \textit{generically trained models}, there are no results listed for some sequences, because some frames of those sequences were part of the training data and therefore the cited publications did not include results for them.

 In Table \ref{table:abl}, we compare four different configurations of SeqNOC where the first configuration is called SeqNOC (baseline) and the other three are obtained by changing only one aspect with respect to the baseline. Additionally, the results for fNNOC \cite{nnoc} are also provided in the last column (since we had the running time results available). In the variant named SeqNOC-LM, a CNN model with about one third of the original number of weights is utilized by reducing the number of output channels in the first 3 layers to 20, 20 and 40; respectively. This improves the encoding and decoding speed whereas it worsens the bitrates for most sequences. In the variant named SeqNOC-10, the number of training frames is chosen to be 10, instead of 5 in the baseline SeqNOC. Having more training frames improves the bitrates whereas the encoding time is noticeably longer since both the collection of training data and the optimization of CNN takes longer. Finally, the variant SeqNOC-SP encodes in a single phase instead of 4 phases in SeqNOC baseline, resulting in best speed whereas the bitrates are a bit higher. The baseline model is observed to provide a reasonable trade-off between speed and bitrate. The addition of the phase-based processing achieves savings in the bitrate (SeqNOC-SP vs. SeqNOC (baseline)) in average about 7\% for the 8i-dataset whereas for the MVUB dataset the improvement is less significant (about 1\%).

The proposed single phase encoding scheme SeqNOC-SP, which encodes a $2\times2$ block of voxels at each AE operation, proves to achieve savings in bitrate between 4\% and 12\% when compared to fNNOC \cite{nnoc}, which encodes only one voxel at each AE operation, but otherwise has a similar context to SeqNOC-SP. This shows that resorting to encoding $2\times2$ blocks of voxels instead of a single voxel leads to a high improvement of the  bitrates, although it makes the implementation slightly more difficult.
  
Furthermore, in Table \ref{table:abl} one can see that the implementation of our solution uses informative contexts in an efficient way, achieving encoding/decoding times between 3 and 10 seconds per each point cloud, at compression rates better than the earlier fast NN solutions fNNOC \cite{nnoc} and MSVDNN \cite{msvoxeldnn}, and at speeds more than 10 times faster. The only solutions having better compression results are achieved by NNOC \cite{nnoc} and VoxelDNN \cite{nguyen2021learning}, (see bitrates in Table \ref{table:compares4d}),  which are more than 100 times slower at decoding than our proposed methods.
 
The implementation of the method will be made public, in case the paper is accepted for publication.

\section{Conclusion}
We have introduced a lossless geometry encoding scheme for sequences of point clouds using CNN models that are designed in a new paradigm, named {\it specifically trained models}, which was not used until now. The learning of the CNN model can be done fast enough at the encoder, so that the learning plus sequence encoding time, divided by the number of frames, gives very competitive per frame encoding times, at a bitrate performance better than that obtained in the competing paradigm, of  {\it generically trained models}. The coding probability models at each 2x2 block, are computed in parallel by the convolutional neural network, at each section $z=z_0$ through the point cloud, contributing to a fast encoding and decoding performance. In order to improve the decoding time, which has unreasonably large values for the published solutions \cite{nnoc,nguyen2021learning,msvoxeldnn}, we adopt a four phase encoding at each section, such that the content of the contexts improves from one phase to another, including the recently encoded/decoded voxels inside the context after each phase. Several variations of the method were proposed, covering various interesting trade-offs (compression ratio vs. time complexity). We discussed the performance of the proposed solution compared to the recently published schemes and found that the introduced features are producing significant improvements.

%\section{Conclusion}
%A conclusion section is not required. Although a conclusion may review the 
%main points of the paper, do not replicate the abstract as the conclusion. A 
%conclusion might elaborate on the importance of the work or suggest 
%applications and extensions. 

%\appendices

%Appendixes, if needed, appear before the acknowledgment.

%\section*{References and Footnotes}

%\reftitle{References}
%\externalbibliography{yes}

%\bibliographystyle{IEEEbib}
\nocite{*}
\bibliographystyle{unsrt}
\bibliography{refs}

\end{document}